# 베트남어 사전을 사용한 베트남어 SentiWordNet 구축

뷔쉬엔손, 박성배
경북대학교 컴퓨터학부
e-mail : {sonvx, sbpark}@sejong.knu.ac.kr

# Construction of Vietnamese SentiWordNet by using Vietnamese Dictionary

Xuan-Son Vu, Seong-Bae Park
Dept. of Computer Science and Engineering, Kyungpook National University

### Abstract

SentiWordNet is an important lexical resource supporting sentiment analysis in opinion mining applications. In this paper, we propose a novel approach to construct a Vietnamese SentiWordNet (VSWN). SentiWordNet is typically generated from WordNet in which each synset has numerical scores to indicate its opinion polarities. Many previous studies obtained these scores by applying a machine learning method to WordNet. However, Vietnamese WordNet is not available unfortunately by the time of this paper. Therefore, we propose a method to construct VSWN from a Vietnamese dictionary, not from WordNet. We show the effectiveness of the proposed method by generating a VSWN with 39,561 synsets automatically. The method is experimentally tested with 266 synsets with aspect of positivity and negativity. It attains a competitive result compared with English SentiWordNet that is 0.066 and 0.052 differences for positivity and negativity sets respectively.

## 1. Introduction

The English SentiWordNet (ESWN) introduced by Esuli et al. [1], is a lexical resource publicly available for research purposes. It has played a vital role in opinion mining or other sentiment analysis applications. Baccianella et al. [2] reported that ESWN has licensed to more than 300 research groups and been used in a variety of research projects worldwide. Thanks to the various available English resources including English WordNet[1], ESWN, subjectivity word list (Wilson et al. [3]), and the like, hence most studies in sentiment analysis domain go with English and leave other languages behind. In an effort to minimize this gap, along with an ambition to bring the same success of ESWN to other languages, we contribute a strategy to generate Vietnamese SentiwordNet (VSWN) with less available of Vietnamese lexical resources. SentiWordNet is essential resource but building it is a difficult task. As our best knowledge, there are few researches work on this task. For English we have first introduction to ESWN by Esuli et al. [1], and improved versions [4; 5; 2]. Das et al. [6], presented SentiWordNet for Indian language, and one more contribution toward to building global SentiWordNet [7].

Due to the unavailability of human labor, both ESWN and ISWN are generated automatically. Firstly, ESWN is generated from English WordNet, then applying machine learning method [4; 2] to calculate sentiment polarities for each synset. Moreover, by taking advantageous of WordNet, random-walk step method [2] on WordNet can improve resultant SentiWordNet considerably. Secondly, ISWN is generated by using machine translator to translate synsets in ESWN to Indian and directly copy sentiment scores for synsets. The approach [2] is state-of-the-art for English while [6] is a very first approach for minor languages with absence of WordNet. Since the latter depends too much on machine translator, therefore we propose to inherit the advantages of machine learning method of the former.

Based on the above investigation, in this work, we adapt machine learning method to generate VSWN. Because Vietnamese WordNet is unavailable, we use Vietnamese dictionary (Vdict) to replace WordNet with respect to extract partial components for each synset in SentiWordNet. Noticeably, there is no information about opinion-related scores in WordNet or in Vdict. Therefore, machine learning method [2] is adapted to calculate those scores. We empirically analyze this approach and present experimental results with aspect of classification performance and resultant VSWN's accuracy. Firstly, to evaluate classification performance, we apply n-fold cross validation on development data set to get the best learning parameters, then applying those parameters on testing data set to get final evaluation results. Secondly, to measure VSWN, a gold standard sentiment word list is constructed to measure the distance from our VSWN to the gold standard.

The rest of the paper is laid out as follows. Section 2 introduces to build ESWN, the method ISWN, and Vietnamese dictionary structures. Section 3 addresses our method to generate VSWN by using Vdict. Evaluation method and results are shown in Section 4. Section 5 gives conclusion and our future studies.

## 2. Related Work

In this section, we will show here the structures of SentiWordNet, current researches about SentiwordNet.

---
[1] www.wordnet.princeton.edu

Additionally, we will describe the structures of Vdict, a part of VLSP Project which will be employed to replace WordNet in this work.

## 2.1 SentiWordNet

SentiWordNet is first introduced by Esuli et al. with ESWN 1.0 [1] that explicitly devised for supporting opinion mining. ESWN is annotated from WordNet by adding the notions of "positivity", "negativity", and "neutrality" for each synset automatically. In SentiWordNet, synset is a set of synonymous terms. A synset is defined as a tuple contains 6 fields including POS (part of speech), ID (synset identifier), PosScore (positive score), NegScore (negative score), Term (a word representing a concept), Gloss (a definition about the term and a sample sentence using term) which are delimited by a <tab>. Though the neutral score of a synset $s$ is not mentioned, it can be easily inferred from the information of a synset by the equation:

$$Neu(s) = 1 - (Pos(s) + Neg(s))$$

In general, there are three approaches including (1) using WordNet [1; 4; 5; 2], (2) using machine translation [6; 7], and (3) using human labor [7]. The first approach takes advantages of WordNet to generate SentiWordNet, while the second translates existing SentiWordNet from one language to other languages. The last approach based on human labor on the Internet to vote sentiment status for each synset. Currently, the method comes from Baccianella et al. [2] is state-of-the-art for building SentiWordNet. Thanks to advantageous of WordNet, most of method have been optimized SentiWordNet based on WordNet. Evidently, WordNet is used to expand training sets with different "radius" k to take the advantages of committee ternary classifiers [1; 2]. By using different radius $k \in \{0, 2, 4, 6\}$ and learning technology (Rocchio and SVM), the committee consists of 8 members. Then, the value of a given synset is calculated as its average Pos (resp., Neg, Obj) value across the committee. Moreover, the random-walk step on WordNet can improve 17.11%, and 19.23% for the ranking by positivity, and the ranking by negativity, respectively. The mathematics behind the random-walk are deeply presented by Esuli and Sebastiani [4]. With the second approach Das et al. fully depends on machine translation to generate SentiWordNet for new languages [6; 7]. This method has two major disadvantages: (1) quality of machine translation is limited, therefore, there may be had a big gap between SentiWordNet and reality data, and (2) all the opinion-related scores are directly copied from source SentiWordNet into destiny SentiWordNet even if each language has its own idiosyncratic sentiment meanings.

In ESWN 3.0, Baccianella et al. [2] used Kendall tau distance (p) into the evaluation step. In ISWN, Das et al. [6] used ISWN to do opinion mining task on the basis of coverage and polarity scores for the evaluation step.

## 2.2 Vietnamese Dictionary

As illustrated from beginning, available lexical resources for Vietnamese are limited except Vdict which is useful to generate VSWN. Vdict is a well form constructed dictionary by using XML language. It consists of 39,561 terms in which each term has essential information about morphology, syntactic, and semantic that are sufficient for deriving synset information. The figure 1 contains sematic information of the term *đẹp* "*beautiful*". The semantic segment contains information about synonym *xinh* "*pretty*", antonym *xấu* "*ugly*", definition *có hình thức hoặc bản chất đặc biệt, làm cho người ta thích nhìn ngắm hoặc kính nể* "*having an attractive appearance or specific characteristics that make other people want to look at or respect*", and particular using context of the term e.g., *phong cảnh đẹp* "*beautiful landscape*".

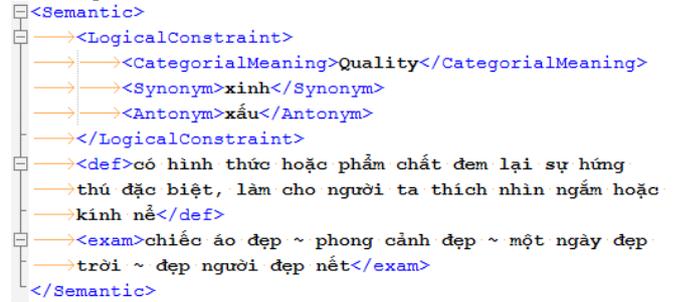

Figure 1: Semantic information of the term *đẹp* "*beautiful*"

## 3. Generating VSWN

After investigation into the construction of ESWN 3.0, ISWN, connecting with the available resources for Vietnamese, we consulted to the following judgements: (1) the method to generate ESWN 3.0 is a prominent method when the WordNet is available, and (2) the method of building ISWN is appropriate only with building the initial SentiWordNet due to it totally depends on machine translation. But, to apply method of ESWN 3.0, there are major problems: (1) there are no available seed sets as Turney and Littman [9] (TLSeeds) contributed in Vietnamese, and (2) WordNet for Vietnamese is unavailable by the time of this paper. Eventually, we propose to use translation method to build VSWN core to replace TLSeeds, and then using machine learning method to generate VSWN from Vdict. Here we describe in details about our approach to construct VSWN.

### 3.1 Framework overview

The overview of our approaching method is illustrated in the Figure 2. There are two main phases in our approach called (1) building VSWN core phase and (2) semi-supervised learning phase. The objectiveness of the first phase is to obtain VSWN core from ESWN. In the second phase, by using the VSWN core, we construct two classifiers to calculate opinion-related properties for all the synsets. The more details about those two phases are deeply described in the following subsections.

### 3.2 Building VSWN Core

With the lacking of Vietnamese lexical resources, we contributed a VNComments corpus to use in building the core. The VNComments corpus contains user's comments crawled from top electric news[2] in Vietnam. After removing noisy data, the corpus consists of 1,596 sentences annotated manually by our annotators by tagging final opinion of each sentence with one of Pos tag, Neg tag, or Neu tag. To guarantee the VSWN core is useful for further processes, we limited the core with following constraints: (1) there is no second sense of a term in the core, and (2) limited domain of terms in Vietnamese news. The first rule letting our annotators work more accurately, while the second rule allowing us to re-calculate opinion-related properties in the second step by using the VNComments corpus.

---

[2] www.vnexpress.net

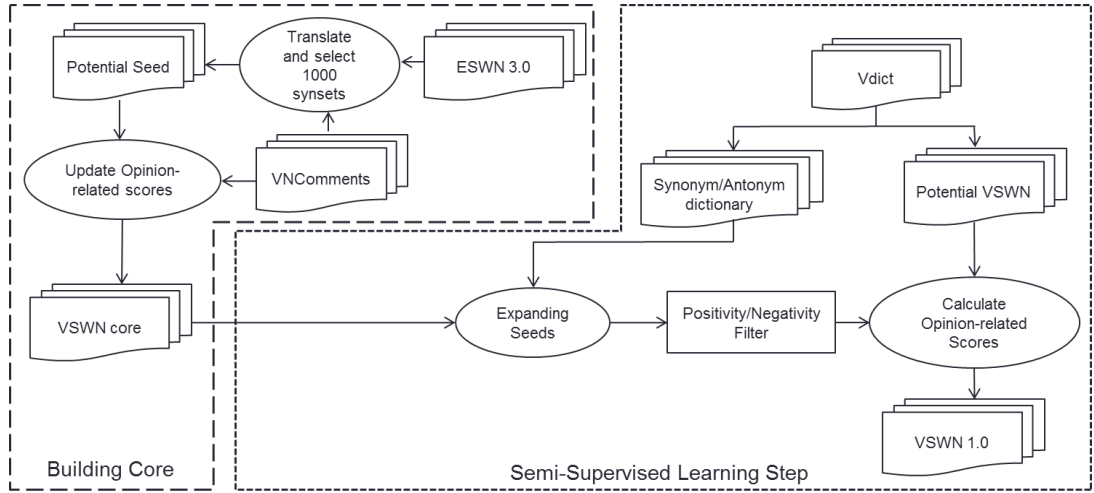

Figure 2: Framework Overview

The building core process consists of two sub-steps: (1) Synsets selection, (2) Updating opinion-related properties.

1. In step (1), we use Google Translate API to translate all the synsets in ESWN which have Pos(s) or Neg(s) above 0.4 in order to get terms that have less ambiguity opinion. The threshold 0.4 is selected based on the heuristic threshold of examining translated results from English to Vietnamese. In fact that the translation quality of Google Translate is limited, then we select coincide terms between VNComments and the translated synsets. Finally, five annotators are employed to select the top 1017 synsets.

2. In step (2), VNComments corpus is employed to recalculate Pos(s), Neg(s) of each term in all the synsets by using following equation:

$$Pos(w) = \frac{\#_{pos}(w)}{\#(w)}$$
$$Neg(w) = \frac{\#_{neg}(w)}{\#(w)}$$

where $\#_{pos}(w)$ is the occurrences of word w in pos set, $\#_{neg}(w)$ is the occurrences of word w in neg set, and $\#(w)$ is the total occurrences of word $w$ in the VNComments corpus.

Eventually, the core has 1,017 synsets containing 1206 terms in which 1,093 terms are adjective and 113 remaining terms are verb. Since 2010 we finished to construct the VSWN core, then it has been proved its benefit when Tien-Thanh et al. [8] used the core to extract sentiment features in customer reviews with F1 is 90%.

**3.3 The semi-supervised learning step**

Since the significant change of ESWN 3.0 is the random-walk step on the English WordNet, that could improve the ESWN 3.0 remarkably compared with ESWN 1.0. With the fact WordNet for Vietnamese is unavailable, therefore in this study, we concentrate on semi-supervised learning step to achieve efficient SentiWordNet.

Based on the process Esuli et al. [1] proposed in the semi-supervised learning step, connected with our circumstances, we propose the following changes: (1) VSWN core is employed to generate the three seed sets instead of WordNet by using a heuristic threshold of 0.3, (2) all entries in Vdict are classified instead of WordNet, and (3) we change to use only SVM which is optimized to achieve highest performance for classification task. Overall the semi-supervised learning process consists of four substeps namely (1) expanding VSWN core, (2) classifier training, (3) synset extraction, and (4) calculate opinion related scores for each synset. The steps are described as follows.

1. In step (1), we divide VSWN core into three subsets: positive set ($Tr_{pos}$), negative set ($Tr_{neg}$), and neutral set ($Tr_{neu}$), in which positive (resp., negative) set is selected from all synsets that have positive (resp., negative) score higher than the threshold 0.3 which is selected after examining the VSWN core. The remaining synsets would be put into neutral set. Additionally, since the amount of VSWN core is limited (1017 synsets), therefore we using synonym (antonym) relation in Vdict to expand the three sets by using following rules: (1) adding to $Tr_{pos}$ (resp., $Tr_{neg}$) all entries having a link to $Tr_{pos}$ (resp., $Tr_{neg}$) based on synonym relation, (2) adding to $Tr_{pos}$ (resp., $Tr_{neg}$) all entries having a link to $Tr_{neg}$ (resp., $Tr_{pos}$) based on antonym relation. To the end of this step, we have the $Tr_{neu}$ set and two expanded sets including $Tr'_{pos}$, $Tr'_{neg}$.

2. In step (2), positivity and negativity classifiers are generated by using combination of $Tr'_{pos}$, $Tr'_{neg}$ and $Tr_{neu}$ as follow: (1) $Tr'_{neg}$ combines with $Tr_{neu}$ to be negative training set of positivity classifier. $Tr'_{pos}$ itself is positive training set for positivity classifier. (2) $Tr'_{neg}$ is positive training set for negativity classifier. $Tr'_{pos}$ combines with $Tr_{neu}$ to be negative training set of negativity classifier. Since the glosses of the synset (resp., entry in Vdict) contain using context of term in synset, therefore glosses are used by training module instead of synset (resp., entry) themselves.

3. In step (3), we extract all entries in Vdict (including those in the VSWN core in the step (1)) to have a potential VSWN. Then those synsets are classified throw two classifiers (positivity classifier and negativity classifier) achieved in step (2).

4. In step (4), each synset is classified via two classifiers

will have two margins represented positive score and negative score. To normalize those scores into the interval [0, 1], we recalculate the scores for each synset as the following equation:

$$f(\phi_i) = \log\left(\frac{\phi_i}{Max\left(abs(\phi_p) + abs(\phi_n)\right)}\right)$$

Where:
(a) $\phi_i$ is a margin of synset
(b) $\phi_p$ is the positive margin of synset
(c) $\phi_n$ is the negative margin of synset

After this step, one synset will achieve positive score and negative score.

## 4. Experiment

In order to get the most effective SentiWordNet, the two classifiers must have the highest performances. Thus we employ various experiments to evaluate and select the most advantageous parameters to perform opinion-related scores calculation task.

### 4.1 Dataset

From the expanding VSWN core step described in 3.3, we achieved the training data consist of 847 positive glosses, 449 neutral glosses, and 1232 negative glosses. Those data is divided into three sets including (1) training set, (2) development set, and (3) testing set by the proportion of 3:1:1, respectively. In more details, we perform 4-fold validation on training set and development set to select the most proper parameters, then using those parameters to apply on the testing set to get the final classification performance. Since positive classifier and negative classifier are using the same combination method from the same training data. Thus we only measure classification performance of positive classifier thanks to the performance of negative classifier would be the same or slightly change but not too different from positive classifier.

As demonstrated from beginning, with lacking of Vietnamese resources, a standard sentiment word list for Vietnamese is unavailable. Thus we manually contributed a gold standard sentiment word list containing 266 synsets for evaluation.

### 4.2 Evaluate classifiers

In table 1 is classification performance in term of F1 values of the positivity classifier with different values of $\gamma$ in SVM for token based and word based model. The table shows that Word Based Model with $\gamma = 1$ achieves the best performance. Therefore these options are selected to train the positivity (resp., negativity) classifier.

|  | $\gamma = 0$ | $\gamma = 1$ |
|---|---|---|
| Token Based | 0.518 | 0.535 |
| Word Based | 0.523 | **0.545** |

Table 1: Classification performance

### 4.3 Evaluate VSWN

After investigation into the both evaluation methods of building ESWN and building ISWN, we found that the former is better than the latter on behalf of one reason is that. The evaluation step should focus only on how well is the SentiWordNet rather than using application level to do evaluation task. Therefor we decided to use a gold standard to measure our contributed SentiWordNet. In evaluation step, Kanddall tau ($\tau_p$) distance is employed to measure the resultant VSWN. The table 2 reports the values for the positivity and negativity rankings as measured on the standard word list. The comparison between the SentiWordNet of different languages seem bias but it partially reflects the reliability of VSWN.

|  | Positivity | Negativity |
|---|---|---|
| VSWN 1.0 | 0.297 | 0.283 |
| ESWN 3.0 | 0.281 | 0.231 |

Table 2: $\tau_p$ values for the positivity and negativity rankings derived from Vdict compared with ESWN 3.0.

## 5. Conclusion

In this paper, we have presented a thorough investigation into SentiWordNet's construction methods for generating VSWN. Our key contributions of the work is three-fold including (1) building a Vietnamese gold standard semantic wordlist for evaluation; (2) generating VSWN without using Vietnamese WordNet; and (3) proving the resultant VSWN is worthy for further study. As the glosses in Vdict are short text, so the performance is suffer from sparsity problem easily. Therefore solving sparsity problem to achieve better performance is our essential further study.

## References

[1] Andrea Esuli and Fabrizia Sebastiani. *Sentiwordnet: A publicly available lexical resource for opinion mining*. LREC - Language Resources and Evaluation (LREC'06), pages 417–422, 2006.

[2] Stefano Baccianella, Andrea Esuli, and Fabrizio Sebastiani. *Sentiwordnet 3.0: An enhanced lexical resource for sentiment analysis and opinion mining*. LREC - Language Resources and Evaluation, 2010.

[3] Andrea Esuli and Fabrizia Sebastiani. *Recognizing contextual polarity in phrase-level sentiment analysis*. North American Chapter of the Association for Computation Linguistics-NAACL, 2005

[4] Andrea Esuli and Fabrizia Sebastiani. *A high-coverage lexical resource for opinion mining*. Technical Report 2007-Tr-02. Istitutiodi Scienza e Technologie dell'Informazione. Consiglio Nazionale delle Ricerche, Pisa, IT, 2007.

[5] Andrea Esuli. *Automatic generation of lexcial resources for opinion mining: Models, algorithms, and application*. Ph.D. thesis, Scuola di Dottorato in Ingegneria "Leonardo da Vinci", University of Pisa, Pisa, IT, 2008.

[6] Amitava Das and Sivaji Bandyopadhyay. *Sentiwordnet for indian languages*. In the 8th Workshop on Asian Language Resources (ALR), 2010.

[7] Amitava Das and Sivaji Bandyopadhyay. *Towards the global sentiwordnet*. Proceedings of the 24th Pacific Asia Conference on Language, Information and Computation, PACLIC 24, Tohoku University, Japan, 2010.

[8] Tien-Thanh Vu, Huyen-Trang Pham, Cong-To Luu, and Quang-Thuy Ha. *A feature based opinion mining model on product reviews in Vietnamese*. Semantic Methods for Knowledge Management and Communication, pages 23–33, 2011.

[9] Christopher D. Manning, Prabhakar Raghavan, and Hinrich Schutze. *An Introduction to Information Retrieval*. Cambridge University Press, Cambridge, UK., 2009.